\documentclass[
twocolumn,
]{ceurart}

\usepackage{listings}
\begin{document}

\copyrightyear{2023}
\copyrightclause{Copyright for this paper by its authors.
  Use permitted under Creative Commons License Attribution 4.0
  International (CC BY 4.0).}

\conference{RecSys in HR'23: The 3rd Workshop on Recommender Systems for Human Resources, in conjunction with the 17th ACM Conference on Recommender Systems, September 18--22, 2023, Singapore, Singapore.}

\title{Enhancing PLM Performance on Labour Market Tasks via Instruction-based Finetuning and Prompt-tuning with Rules}

\author[1]{Jarno Vrolijk}[%
email=j.vrolijk@uva.nl,
]
\author[2]{David Graus}[%
email=david.graus@randstadgroep.nl,
]

\address[1]{University of Amsterdam, Amsterdam, The Netherlands}
\address[2]{Randstad, Diemen, The Netherlands}

\begin{abstract}
The increased digitization of the labour market has given researchers, educators, and companies the means to analyze and better understand the labour market. However, labour market resources, although available in high volumes, tend to be unstructured, and as such, research towards methodologies for the identification, linking, and extraction of entities becomes more and more important. Against the backdrop of this quest for better labour market representations, resource constraints and the unavailability of large-scale annotated data cause a reliance on human domain experts. We demonstrate the effectiveness of prompt-based tuning of pre-trained language models (PLM) in labour market specific applications. Our results indicate that cost-efficient methods such as PTR and instruction tuning without exemplars can significantly increase the performance of PLMs on downstream labour market applications without introducing additional model layers, manual annotations, and data augmentation.
\end{abstract}


\maketitle

\section{Introduction}
The increasing availability of raw labour market information allows businesses, educational facilities and job seekers to gain a clear and more complete understanding of the labour market \cite{khaouja2021}. 
While the increasing volumes of available data provide opportunities, there are several challenges towards fully utilizing the data. 

On the one hand, the majority of available data is of an unstructured nature, with a lack of large-scale annotated datasets that could be used in training, and/or fine tuning of models for downstream applications. 
%
On the other hand, there is much effort in creating structured representations of labour market data, through taxonomies and ontologies such as ESCO,\footnote{\url{https://esco.ec.europa.eu/en}} ISCO,\footnote{\url{https://www.ilo.org/public/english/bureau/stat/isco/isco88/}} or O*NET.\footnote{\url{https://www.onetcenter.org/overview.html}}

Leveraging structured ontologies to enrich and interpret unstructured labour market data has considerable research attention, through skill and occupation recognition, classification, and linking~\cite{zhao2015, sibarani2017, vrolijk2022, zhang2022, zhang2023}. 
These different downstream tasks can prove invaluable in enabling better workforce and labour market insights, identification of trends and temporal patterns, and providing structured data or enrichments that can be applied as feature representation for job or career path recommendations~\cite{degroot2021job,KETHAVARAPU2016915,info13110510}. 

Many of these approaches rely on supervised learning, where a commonly identified limitation in literature is the availability of multilingual, labour market and task-specific datasets. 
In addition, due to the dynamic nature of the labour market makes it very difficult to keep more structured representations of the labour market up-to-date and relevant (i.e. labour market ontologies and taxonomies); updating and maintaining these knowledge structures is typically done by human domain experts, making them time- and resource-intensive, and meaning that whenever such a structure is updated, datasets for supervised learning may become obsolete. 

In this paper, we propose a novel method that relies on pretrained language models (PLMs), prompt tuning with rules (PTR)~\cite{han2022}, and the structured multilingual ESCO taxonomy, to efficiently and cheaply generate large amounts of labeled data for learning a variety of down-stream tasks for extracting structured information from unstructured labour market data, specifically: 
(i) relation classifiers, that aim to predict the type of relation between skills and occupations, 
(ii) entity classifiers, that aim to classify labour market entities as skill or occupation, 
(iii) entity linkers, which aims to link various surface forms of labour market entities to their canonical underlying skill or occupation entity, and 
(iv) question answering approaches, that aim to answer the correspondence between a descriptive text and the associated skill or occupation.

In this paper, we aim to address the following research questions: 
\begin{enumerate}
    \item Are "out-of-the-box" PLMs capable of generalizing learned behavior to labour market specific applications?
    \item Does instruction, and sub-prompt finetuning a PLM on a mixture of task-specific (i.e. general and labour market specific) datasets increase the performance on labour market specific benchmarks?
    \item Is the tuned PLM able to transfer the learned behavior across labour market specific tasks?
\end{enumerate}

In this paper, we demonstrate domain-specific prompt-based tuning and its effect on the performance of skill extraction, occupation classification, link prediction, and entity linking tasks. 
We propose leveraging instruction tuning without exemplars (i.e. no examples at inference time) and sub-prompts for a more cost-efficient solution for downstream labour market applications \cite{chung2022, gao2020, schick2020, han2022}.
We provide manually constructed templates that encode the knowledge embedded in the ESCO occupation and skill taxonomies. 
We benchmark different configurations of finetuning the PLMs, to demonstrate the effectiveness of e.g., adding instructions or sub-prompts. 
\section{Related Work}
Recent successes of PLMs such as GPT~\cite{radford2018}, BERT~\cite{devlin2019}, RoBERTA \cite{liu2019} and T5 \cite{raffel2020} have demonstrated the usefullness and adaptability of the transformer architecture. 
Although these PLMs can capture rich knowledge from massive corpora, a fine-tuning process with extra task-specific data is still required to transfer their knowledge for downstream tasks. 
Besides fine-tuning language models for specific tasks, recent studies have explored better optimization and regularization techniques to improve fine-tuning.

Several works try to integrate ontological and/ or taxonomical knowledge into task-specific models, to improve the performance of downstream applications. Take the work by \cite{peters2019}, who introduced \textit{KnowBert}, a methodology that explicitly models entity spans in the input text. They further use an entity linker to retrieve relevant embeddings of the entity from a knowledge base to enhance their representations. Another approach would be the work by \cite{bordes2013}, the so-called \textit{TransE} model, that focused primarily on representing hierarchical relationships. Similar to our work, \textit{TransE} models multi-relational data from knowledge bases (i.e. triplestores) to improve performance for link prediction \cite{bordes2013}.

\cite{zhang2019ernie} took a different approach, proposing \textit{ERNIE}, a method that consists of two stacked modules, namely the T-Encoder responsible for capturing lexical and syntactic information, and the K-Encoder responsible for augmenting this lexical and syntactical information with extra token-oriented knowledge from the underlying layer \cite{zhang2019ernie}. Lastly, we have \textit{ESCOXLM-R} that employ further pre-training on the ESCO taxonomy \cite{zhang2023}. In addition to the masked language modelling (MLM) pre-training objective, the authors also introduce the so-called ESCO Relation Prediction (ERP) task to internalize knowledge of non-hierarchical relations within ESCO \cite{zhang2023}.

Another pre-training-based approach that leverages a self-supervised method to pre-train a deeply joint language-knowledge foundation model from text and knowledge graphs at scale is the Deep Bidirectional Language-Knowledge Graph Pretraining (DRAGON) proposed by \citet{yasunaga2022b}. 
Results from the paper indicate that DRAGON outperforms existing LM and LM+KG models on diverse downstream tasks in particular on complex reasoning about language and knowledge.

Despite the success of fine-tuning PLMs, there is a big gap between the MLM objective and fine-tuning objectives for downstream applications. Prompt-based learning has been a widely explored method that uses templates to transform the input into classification problems, and as such, closes the gap between task-specific and MLM objectives \cite{liu2023}. \cite{chen2022knowprompt} propose \textit{KnowPrompt}, a method for task-oriented prompt template construction where they use special markers to highlight entity mentions in the template. \cite{cui2021} also proposed a template-based NER model using BART. The model enumerates all possible text spans and considers the generation probability of each type within manually crafted templates \cite{liu2023, cui2021}.

Since the manual creation of templates is labour-intensive, methods for the automated generation of prompts and labels are well-researched. In principle, a prompt consists of a template and label words. As such, \citet{schick2020} first searches the label word space for the manually created templates. Next, gradient-guided search automatically generates both templates and label words. Compared to human-picked prompts, most auto-generated prompts cannot achieve comparable performance \cite{han2022}.

Prior literature has shown that increasing the number of tasks in finetuning improves the generalization to unseen tasks \cite{chung2022}. Experiments from ~\citet{chung2022} show that "instruction finetuning" scales well with the number of tasks and the size of the model. \citet{wei2021} further suggests that instruction-tuned models respond better to continuous outputs from prompt tuning. Prompt tuning on FLAN even achieves more than 10\% improvement over a non-instruction-tuned equivalent model \cite{wei2021}.\par

\section{Methodology}

\begin{figure*}[!t]
    \centering
    \includegraphics[width=\textwidth]{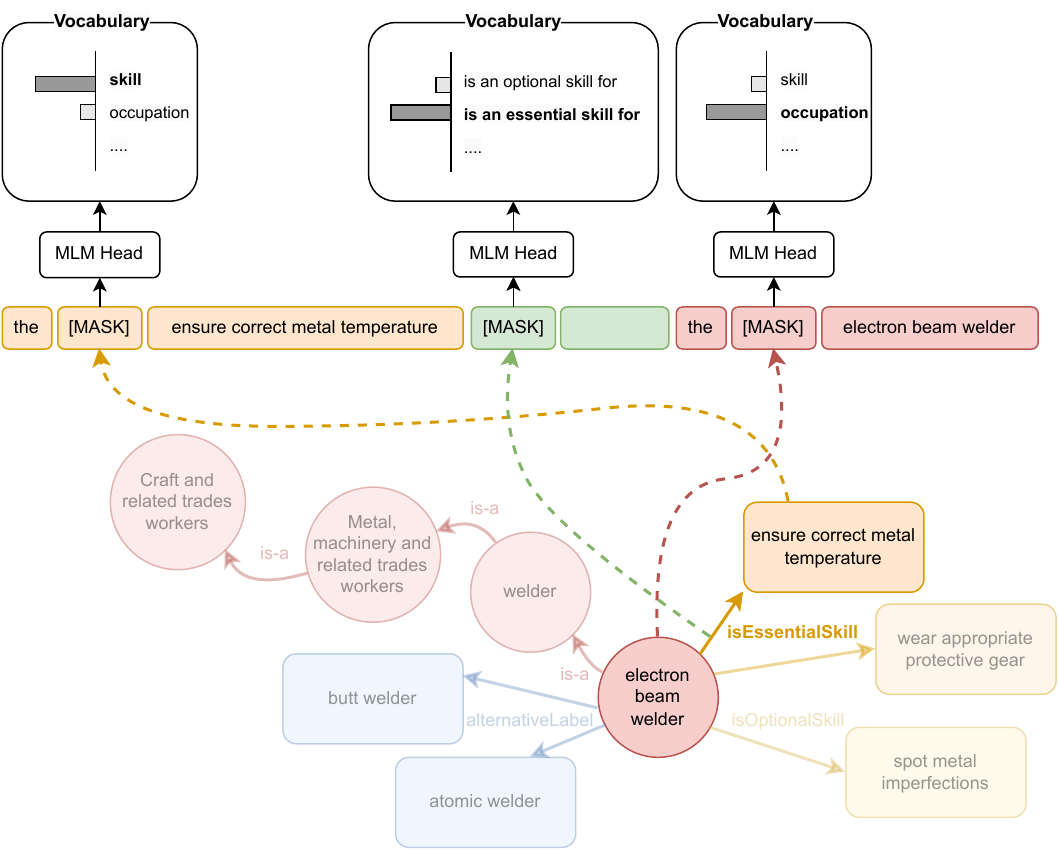}
    \caption{Visual representation of proposed method: PTR (shown on top) with three sub-prompts (yellow, green, and red) with MLM heads predicting [MASK] tokens, given their respective verbalizers (inspired by \citet{han2022}). 
    An outtake of the ESCO taxonomy represented in the bottom, with hierarchical (red) relations and non-hierarchical (rest), and how the entities and relations populate the template (dotted lines).}
    \label{fig:method}
\end{figure*}

\subsection{Preliminaries}

\subsubsection{ESCO}
ESCO (European Skills, Competences, Qualifications and Occupations) is the European multilingual classification of skills, competences and occupations. In total, ESCO describes 3,008 occupations and 13,980 knowledge, skill, and competences in 28 different languages.

ESCO has both hierarchical and non-hierarchical relationships: hierarchical relationships, or hypernymies, are relations of the form \textit{x is-a y}~\cite{roller2018, vrolijk2022}. 
Non-hierarchical relationship are essentially those relationship that are not hierarchical. 
For example: ``\textit{Java Programming} is an essential skill for a \textit{Software Developer}'' is a non-hierarchical relationship, whereas, ``a \textit{Software Developer} is a \textit{Information and Communications Technology Professional}'' is hierarchical.

\subsubsection{PLMs: T5 \& FLAN-T5}
In this paper we rely on the T5 PLM, since the text-to-text framework allows us to directly apply the same model, objective, training procedure, and decoding process to every task we consider \cite{raffel2020}.
In addition, we turn to an instruction-tuned variant of the T5 model: FLAN-T5. 
Instruction-based finetuning has shown to improve zeroshot performance on unseen tasks~\cite{wei2021, chung2022}.
In this paper we aim to study whether this property also applies to the domain-specific unseen tasks that we propose.

\subsection{Prompt-Tuning with Rules}
In this paper, we utilize the ESCO taxonomy as background knowledge for the Prompt-Tuning with Rules (PTR) approach proposed by \citet{han2022}. 

PTR builds on prompt tuning methods that rely on cloze tests, where the PLM is applied to replace or fill in a missing word in a sentence. 
A so-called verbalizer maps a fixed set of class labels (e.g., \texttt{positive}, \texttt{negative}) to underlying label words (e.g., ``great'', ``terrible''), so that by predicting a label word, the PLM effectively classifies a sentence. 

PTR extends this prompt tuning approach with prior knowledge encoding, i.e., leveraging logic rules to encode prior knowledge about tasks and classes into prompt tuning, and efficient prompt design, through composing multiple sub-prompts and combining into prompts~\cite{han2022}. 

\paragraph{Illustrative example}
We illustrate how we leverage the ESCO taxonomy to construct and populate sub-prompts, as proposed by \citet{han2022}. 

Consider a (sub-)prompt template for entity type classification, as; 
\texttt{"[CLS] the [MASK] \textit{[ENTITY]}."}
Which can be instantiated for the skill "ensure correct metal temperature," as:
\texttt{"[CLS] the [MASK] \textit{ensure correct metal temperature},"}
and for the occupation "electron beam welder," as:
\texttt{"[CLS] the [MASK] \textit{electron beam welder}."}

Finally, we can combine the above instantiations of the same sub-prompt into a final prompt, that spans entity type and relation classification, as such: 
\texttt{"[CLS] the [MASK]$_1$ \textit{ensure correct metal temperature} [MASK]$_2$ the [MASK]$_3$ \textit{electron beam welder}"}.

PTR relies on so-called "verbalizers" that map class labels to label words. 
In our example, the class labels \{"skill", "occupation"\} for entity classification are mapped to (the same) label words \{"skill", "occupation"\} in place of [MASK]$_1$ and [MASK]$_3$, and the class label \{"isEssentialSkill", "isOptionalSkill"\} in place of [MASK]$_2$ are mapped to the corresponding label words \{"is an essential skill for", "is an optional skill for"\} in the case of relation classification. 

i.e., $\varphi_{[MASK]_1}$ and $\varphi_{[MASK]_3}$ aim to assign an entity class through predicting a label word from X, and $\varphi_{[MASK]_2}$ aims to classify the type of relation between the two through label words Y.

\subsection{Instruction-based Finetuning}
Instruction-based finetuning aims to teach a PLM to perform certain tasks, by responding to instructions in natural language~\cite{wei2021}. 
For two of our three datasets (i.e., the QA and EL), we manually constructed templates that result in natural language instructions that describe the task for that dataset to the PLM.

While scaling language model sizes seems to be a reliable predictor for improved model performance, it comes at the price of high compute. 
Therefore, development of compute-efficient techniques that improve performance at the cost of a relatively small amount of computational resources is important. 
Instruction-based finetuning improves performance of PLMs on evaluation benchmarks by up to 9.4\%, requiring only 0.2\% of the pre-training compute~\cite{chung2022}. Furthermore, \citet{chung2022} demonstrate that smaller models that are instruction tuned can outperform larger models without it.

\begin{figure}
    \centering
    \resizebox{1\columnwidth}{!}{%
    \includegraphics{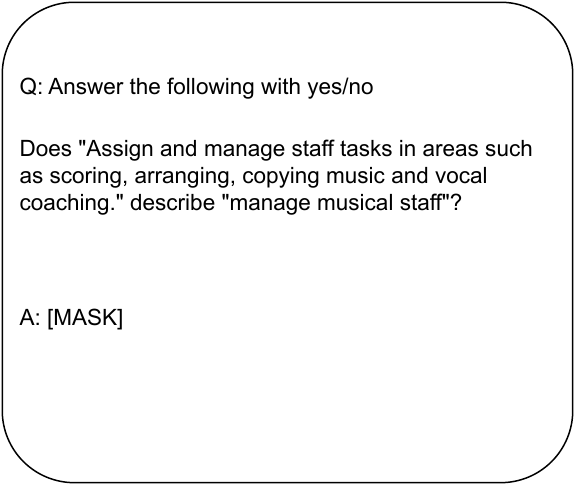}}
    \caption{Visual representation of method: Instruction tuning for the QA examples. The instruction is prepended to the question, instructing the PLM how to proceed in answering the given question.}
    \label{fig:qa-template}
\end{figure}

Figure \ref{fig:qa-template} demonstrates how we leverage the ESCO taxonomy to construct instruction tuning templates for the QA examples. 

\section{Experimental Setup}
The aim of this paper is to leverage prompt-based and instruction-based finetuning, to cost-efficiently optimize PLMs performance on four downstream labour market tasks.
As described in the previous section, we propose four different tasks for evaluation, namely: 
entity classification (EC), 
relation classification (RC), 
entity linking (EL), and 
question answering (QA).

\subsection{Datasets}
We evaluate PTR and instruction-based finetuning in labour market-specific downstream tasks through benchmark datasets we generate through populating hand-crafted templates, with instances from the ESCO taxonomy. 

We generate three datasets of prompts, that address four different tasks; 
i) entity classification (EC) and 
ii) relation classification (RC) as illustrated above (combined in a single set of prompts), 
in addition to 
iii) entity linking (EL), and 
iv) question answering (QA). 

Construction of a self-supervised dataset comprises three different components; 
i) a subset of ESCO relations, 
ii) a template to map the triples associated to the relations to (sub-)prompts, and 
iii) verbalizers that map class labels to label words.

\begin{table}[]
    \centering
    \resizebox{1\columnwidth}{!}{%
\begin{tabular}{lrrr}
\toprule
              & EC + RC & QA    & EL     \\ \midrule
\# total       & 123,752   & 27,792 & 195,350 \\
\# skills      & 13,890    & 13,890 & 13,890  \\
\# occupations & 3,008     & 3,008  & 3,008   \\
\# essential   & 64,877    & -     & -      \\
\# optional    & 58,875    & -     & -      \\
\# altlabels   & -        & -     & 96,117  \\
\# pos         & 123,752   & 13,896 & 97,675  \\
\# neg         & 0        & 13,896 & 97,675  \\ 
\bottomrule
\end{tabular}}
\caption{Statistics of the different datasets. Since, the train and evaluation sets differ due to random sampling or the choice for \textit{K}, we can only report the total counts.}
\end{table}

\subsubsection{Entity Classification + Relation Classification}
To build the entity classification and relation classification (EC + RC) dataset, we leverage the \textit{isEssentialFor} and the \textit{isOptionalFor} relations as found in ESCO.
For both entity classification and relation classification, we largely follow the work by~\citet{han2022}, i.e., we extract all triples that have as subject a skill entity, the \textit{isEssentialFor} or the \textit{isOptionalFor} as predicate, and finally as object an occupation entity.

Formally, our triples look as follows:

\begin{equation}
    <Skill, r ,Occupation>,
\end{equation}
where $r \in\{isOptionalFor, isEssentialFor\}$.
The entity and relation classification template $T(x)$ is formalized as:
\begin{align*}
    s_{1}=The\; [MASK]\; entity\; [Skill] \\
    s_{2}=The\; [MASK]\; entity\; [Occupation] \\
    s_{1}\; [MASK]\; s_{2}
\end{align*}
Lastly, we formulate two different verbalizers $\varphi_{1}$ and $\varphi_{2}$ such that:
\begin{align}
    \varphi_{1}=C_{1}\rightarrow\nu_{1}, \\
    \varphi_{2}=C_{2}\rightarrow\nu_{2},
\end{align}
where $C_{1}$=\{\textit{'Occupation'}, \textit{'Skill'}\} and the accompanying label words $\nu_{1}$ = \{\textit{'occupation'}, \textit{'skill'}\}.
Similarly, $C_{2}$ = \{\textit{'isOptionalFor'},\textit{ 'isEssentialFor'}\}, and the label words $\nu_{2}$ = \{\textit{'is optional for'}, \textit{'is essential for'}\}.

Note that in our case, the verbalizers are one-to-one mappings, whereas in the PTR methodology, many-to-one mappings are also supported. 
For the entity and relation classifications we have not included the possibility of "no relation" and/ or "no entity", for the simple reason of self-supervision. 
While we fully believe these negative examples to be useful for better learning how to recognize entities and the relation connecting entities, they would require manual annotation, and as such fall beyond the scope of this research.

\subsubsection{Entity Linking}
To model entity linking, we rely on the \textit{alternativeLabel} relation in ESCO, i.e., our task is to map an entity surface form or entity mention (\textit{alternative label}), to the canonical entity name (i.e., \textit{label}).

We can formalize the entity linking task as the following triplestore:

\begin{equation}
    <e\;,r\;,m>,
\end{equation}
where $e \in E$ the set of skill and occupation entities, and $m \in M$ the set of skill and occupation mentions (i.e., alternative labels for the ESCO skill and occupation labels). 
Lastly, $r \in C$, meaning that the predicate can be either signalling that the mention is an alternative label or not an alternative for the given ESCO entity. 

Given an entity, $e$ and a mention $m$, we are interested in finding out what type of relation there exists between $e$ and $m$. 
As such, we formalize the entity linking as a masked language problem via $x_{prompt}=e\; [MASK]\;m$.

In Figure~\ref{fig:method}, the blue boxes represent two examples of alternative labels for the occupation \textit{electron beam welder}. 

We formalize the template for our second set of prompts as:
\begin{align}
    \texttt{e [MASK]}\; \nonumber \texttt{m}
\end{align}

We formulate the verbalizer $\varphi$ such that:
\begin{align}
    \varphi = C \rightarrow \nu,
\end{align} 
where 
\begin{align*}
    C=\{\texttt{alternativeLabel}, \texttt{noAlternativeLabel}\} \\
    \nu=\{\texttt{is an synonym for}, \texttt{it not a synonym for}\}
\end{align*}

For each generated example from the ESCO triplestores, we also randomly sample negative examples by randomly shuffling the objects and subject of the positive triplestores and changing the predicate label to \textit{noAlternativeLabel}.

\subsubsection{Question Answering}
For the QA task, we use so-called \emph{instructional templates} as defined by~\citet{chung2022} and \citet{wei2021}. 
Instructional templates prepend an instruction to the prompt. 
In our case, we prepend the example and question with 
\textit{"Answer the following with yes/no"}, instructing the PLM how to answer the question that follows.

The question and answering dataset is constructed with the descriptions of the entities in ESCO. 
As such, we can construct a dataset as: 
\begin{equation}
    <e, \{\texttt{description}\}>
\end{equation}

Next, we define the template $T(x)$ as depicted in the example in Figure \ref{fig:qa-template}. Where we first prepend the instruction \textit{"Q: Answer the following with yes/no"} to the body \textit{"Does [description] describe [entity label]"}, to finish it off with \textit{"A: [MASK]"}.

The verbalizer $\varphi$ then maps the $\{'yes', 'no'\}$ to the label words $\{'yes', 'no'\}$.

We randomly sample correct examples, in addition to generating negative examples by randomly sampling a skill or occupation entity, and pairing this with a randomly sampled description from the set of available descriptions, and tagging the label for the answer to be \textit{"no"}.
This results in a balanced dataset, with a fifty-fifty split of positive and negative examples.

\subsection{Experiments}
\label{subsec:experiments}
In order to answer our research questions, we propose the following experiments.

\subsubsection{Experiment 1: Zero-shot Learning}
First, to better understand the labour market-specific tasks that we propose, we first test off-the-shelve PLMs in a zero-shot setting, using our own generated prompt datasets for inference. 

In addition, to test the hypothesis that FLAN-T5's multitask learning enables a better ability of learning additional (domain-specific) tasks, in our first experiment we directly compare off-the-shelve T5 and FLAN-T5 models, on each of our three datasets.

\subsubsection{Experiment 2: K-shot Learning}
Next, having established the performance differences between the off-the-shelve PLMs, we study the impact of few-shot learning to steer the best performing PLM from experiment 1 towards the domain-specific data and tasks, where we perform an ablation study on the number of examples ($K$) we use for few-shot learning. 

This is motivated by a.o., \citet{han2022}, who report comparable or even better results in the few-shot scenario than e.g., methods that inject special symbols to index the positions of entities and methods that inject both type information and special symbols. 
The authors sample $K$ training instances and $K$ validation instances per class from the original training set and development set, and evaluate the models on the original test set. 

We propose using $K=\{64, 128, 256\}$ sets.

\subsubsection{Experiment 3: Multitask Learning}
After having studied the effect of few-shot learning, we perform an ablation study to measure the effect of learning multiple tasks in parallel, i.e., transfer learning from one task to the other. 

We do so by fine-tuning the FLAN-T5 on all combinations of tasks from a single to the full set, i.e., we train FLAN-T5 on the RC+EC and consecutively on the EL and QA tasks. 
We then test the performance of the resulting model on all three data sets to identify whether, e.g., prompt tuning on EL can help performance on the QA dataset.

\subsection{Implementation Details}

Our model implementation relies on the HuggingFace, PyTorch and OpenPrompt frameworks (albeit with some customizations), proposed by \citet{wolf2020, paszke2019} and \citet{ding2021} respectively. 

For the zero-shot approach of the first experiment, we turn to T5 and FLAN-T5, for which we use the implementation by the original authors \cite{raffel2020, wei2021, chung2022}. 
More specifically, we use the 3 billion parameter checkpoints as found on huggingface under the names; \textit{'t5-3b'}, and \textit{'google/flan-t5-xl'}. 

To answer our second research question we adjust the number of examples used for training the models by comparing different values for parameter $K$ (i.e., number of samples). 
We optimize our PTR and Instruction-based finetuning models using AdamW, with a learning rate of respectively $3e-5$ and $2e-5$. 
Furthermore, we reset the weight decay on the normalization layers and bias. 
We fine-tune all models using batch size 32, and train the PTR models for 10 epochs, whereas, we train the instruction based finetuning models for only 5. 
The best model checkpoint is selected.

\subsection{Evaluation metrics}
In order to systematically evaluate few-shot performance, we randomly pick \textit{K} samples from the total dataset, and use the remaining data to sample evaluation sets. 
This sampling is done 9 times, each iteration we sample 512 random examples from the remaining data after the train/ test split. 
We report F1 scores averaged over 9 runs in addition to standard deviations ($^{\pm std}$). 
We argue that sampling multiple splits gives a more robust measure of the actual performance.

Since the single EC+RC dataset contains two separate tasks, it is important to avoid contamination between the train and test sets. 
Therefore, after the initial division, we check all individual skill and occupation entities from the train set, and remove all relations in the test set that contain any of those entities. 
For the QA and EL training data the risk of contamination is mitigated through the train/test split (i.e., after the normal split unique entries belong either to the train or test set). 

\section{Results}
In this section we present and summarize the results of our experiments described in Section~\ref{subsec:experiments}.

\subsection{Experiment 1: Zero-shot learning}
See Table~\ref{tab:exp1} for the comparison of T5 and FLAN-T5 in 0-shot learning, i.e., applied off the shelve for inference on our generated prompts. 

First, we see that FLAN-T5 substantially outperforms the non instruction-based finetuned counterpart T5 on the QA (83.44 vs. 33.75 respectively) and EL tasks (57.38 vs. 33.89 respectively), but slightly underperforms on the EC+RC tasks, at 44.54 for FLAN-T5 and 48.07 for the T5 model. 

A potential explanation for this might be the fact that FLAN-T5 is trained on a variety of entity classification tasks that do not involve skill and occupation entities (i.e., the primary focus is on person and organisation entities). As such, the learned patterns may interfere with the PLMs ability to recognize skills and occupations. 

\begin{table}[]
    \centering
    \resizebox{1\columnwidth}{!}{%
    \begin{tabular}{llll}
    \toprule
    Model   & EC+RC & QA    & EL    \\ 
    \midrule
    T5      & 48.07$^{\pm.19}$    & 33.75$^{\pm.2}$ & 33.89$^{\pm.37}$ \\
    FLAN-T5 & 44.54$^{\pm.66}$    & 83.44$^{\pm.44}$ & 57.38$^{\pm.60}$ \\
    \bottomrule
    \end{tabular}
}
\caption{F1 scores of experiment 1, where we compare 0-shot performance between T5 and FLAN-T5.}
\label{tab:exp1}
\end{table}

\subsection{Experiment 2: K-shot learning}
In Table~\ref{tab:exp2} we show the performance differences at different levels of $k$ in the fewshot learning scenario. 

First, we note how the performance of FLAN-T5 + PTR substantially outperforms both T5 and FLAN-T5 from Table~\ref{tab:exp1} with F1 scores between 50.42 and 51.60 across different values of $K$, compared to 48.07 and 44.54 respectively for the zero-shot T5 and FLAN-T5. 

Next, we see that different values of $K$ are optimal for different tasks; with maximum scores at $K=128$ for EC+RC and QA at 51.60 and 94.23 respectively, and a maximum score of 98.06 for $K=256$ for EL. 

The scaling of the model potentially gives us insights into how sample efficient the model is in learning the behavior. Larger models are in-general more sample efficient and as such require less examples to learn a particular behavior \cite{liu2022sample}.

\begin{table*}[]
    \centering
    \resizebox{1\textwidth}{!}{%
\begin{tabular}{llllllllll}
\toprule
          & \multicolumn{3}{c}{EC+RC}                                               & \multicolumn{3}{c}{QA}                                                     & \multicolumn{3}{c}{EL}                                                     \\
             Model $\downarrow$/$K$ $\rightarrow$  & \multicolumn{1}{c}{64} & \multicolumn{1}{c}{128} & \multicolumn{1}{c}{256} & \multicolumn{1}{c}{64} & \multicolumn{1}{c}{128} & \multicolumn{1}{c}{256} & \multicolumn{1}{c}{64} & \multicolumn{1}{c}{128} & \multicolumn{1}{c}{256} \\ 
               \midrule
FLAN-T5 + PTR    & 50.42                  & 51.60                   & 50.87                         & -                      & -                       & -                       & -                      & -                       & -                       \\
FLAN-T5 + Instruction tuning & -                      & -                       & -                       & 92.09                  & 94.23                   & 93.71                   & 89.26                  & 95.17                   & 98.06                   \\ 
\bottomrule
\end{tabular}
}
\caption{F1 scores for experiment 2, comparing the impact of number of instructions (e.g., $K$) across the three benchmark datasets (top row).}
\label{tab:exp2}
\end{table*}

\subsection{Experiment 3: Multitask learning}
Finally, we show the impact of learning single or multiple tasks at once. 
Results of our ablation experiments are shown in Table~\ref{tab:exp3}, where we vary with models that are trained on all combinations of different train sets of prompts, which we evaluate on each of the three test set of prompts.

Here, we note that first, in some cases adding prompts for additional tasks increases performance for the original tasks, consider, e.g., the case for (testing on) EL, where adding QA prompts yields an F1-score of 97.61 (row 4, Table~\ref{tab:exp3}), and adding EC+RC prompts even gets performance up to 98.48 (row 5, Table~\ref{tab:exp3}), whereas the model tuned with EL prompts only, scores 95.17 F1 (row 3, Table~\ref{tab:exp3}). 

However, this does not apply for QA nor EC+RC, where only tuning with respectively QA and EC+RC prompts yields the highest score, nor for the case of training on all additional prompts---these runs (bottom row in Table~\ref{tab:exp3}) do not outperform the best performing models tuned on one or two sets of prompts. 

Overall, this indicates that multitask learning can contribute in some cases to increased performance.

\subsubsection{Unseen task performance}
Supporting these observations is the pattern around performance on unseen tasks, i.e., models tuned on (a) task(s) that do not include the test task used for evaluation. 
Consider, e.g., EL; models that have not seen any EL prompts in their tuning stage, perform substantially worse with 58.96 for EC+RC and QA, 57.40 for EC+RC, and 60.31 for QA, versus between 95.17 and 98.48 for models that have seen EL prompts. 

Similar patterns are seen with the other tasks, where for EC+RC models that have not seen any EC+RC prompts perform between 45.05--47.68, and around 50.55 and 51.60 for models that have. 
For QA, we see that models without QA prompts in tuning score between 68.22--87.98, and models that have range from 93.24 to 94.23. 

However, increasing the number of tasks in tuning does increase performance for unseen tasks in two out of three cases: when testing on the EC+RC prompts, a model that combines QA and EL prompts in tuning scores 47.68, and outperforms QA-only (45.93) and EL-only (45.04) models. 
Similarly, for QA, combining EC+RC and EL prompts yields an F1-score of 87.98, whereas EC+RC-only yields 78.36, and EL-only a mere 68.22 F1. 

Finally, models that are tuned on all tasks do not outperform models tuned on two tasks in two out of three sets (only for QA does the full model perform better than models trained on two tasks).

  \begin{table}[]
    \centering
    \resizebox{1\columnwidth}{!}{%
    \begin{tabular}{llll}
    \toprule
    Train $\downarrow$ / Test $\rightarrow$ & \multicolumn{1}{c}{EC+RC} & \multicolumn{1}{c}{QA} & \multicolumn{1}{c}{EL} \\ 
    \midrule
        EC+RC    & \textbf{51.60}$^{\pm.47}$           & 78.36$^{\pm.86}$                  & 57.40$^{\pm.16}$         \\
        QA       & 45.93$^{\pm.26}$               & \textbf{94.23}$^{\pm.24}$              & 60.31$^{\pm.12}$         \\
        EL       & 45.04$^{\pm.26}$               & 68.22$^{\pm.70}$                  & 95.17$^{\pm.39}$         \\
        \midrule
        EC+RC, QA    & 51.34$^{\pm.23}$               & 93.24$^{\pm.21}$                  & \textit{58.96}$^{\pm.52}$         \\
        EC+RC, EL    & 51.21$^{\pm.45}$                        & \textit{87.98}$^{\pm.31}$                  & \textbf{98.48}$^{\pm.29}$         \\
        QA, EL        & \textit{47.68}$^{\pm.23}$                        & 93.69$^{\pm.24}$                  & 97.61$^{\pm.14}$                  \\
        \midrule
        all & 50.55$^{\pm.70}$                        & 94.10$^{\pm.27}$         & 98.19$^{\pm.32}$                  \\ 
    \bottomrule
    \end{tabular}
    }
    \caption{F1 scores of our previously best performing model: FLAN-T5 with 128-shot learning, on the different combinations of tasks we propose.}
    \label{tab:exp3}
\end{table}

\section{Discussion}
Our paper explored three different questions. First, \textit{are "out-of-the-box" PLMs capable of generalizing learned behavior to labour market specific applications?} In order to answer this question, we created three self-supervised benchmarks from the ESCO taxonomy. 

To answer this question, we performed zero-shot comparing between T5 and the instruction-tuned FLAN-T5, that has seen 1,836 additional tasks in prompts. 
Results showed that FLAN-T5 substantially outperforms T5 on two labour market-specific tasks, with a 49.7\% increase in F1 score for QA, and 23.5\% for EL. 
However in the EC+RC task where T5 outperforms FLAN-T5 by 3.53\%. 
These findings confirm that overall, the instruction-tuned FLAN PLM benefits from having seen multiple tasks. 
The result for the EC+RC task can be explained by "misleading" patterns learned from the more general finetuning on named entity recognition (i.e., recognition of "Persons" and "Organizations", etc.). However, further investigations and ablation studies on general task tuning and its exact influence on the performance is needed for a more definite answer.

On the second question, whether \textit{instruction and/or sub-prompt finetuning a PLM on a mixture of task-specific datasets could increase the performance on labour market specific benchmarks?}, we performed experiment 2, where we varied our $K$ instruction samples for training our best-performing PLM: FLAN-T5. 
Results demonstrated that PTR-based finetuning with 128 examples leveraged the best performance. 
Overall, this yielded an 7.06\% performance increase over the zero-shot performance of FLAN-T5. 
Additionally, further scaling of the number of examples, to 256, yielded only a 6.24\% increase, suggesting no further performance increases for further scaling of the number of examples. 
Our results seem to indicate that using PTR with labour market specific examples yields improvements above and beyond the 1836 tasks FLAN-T5 was tuned on.

Lastly, we investigated \textit{the effects of transfer learning across labour market specific tasks.} 

Here, our results suggest that first, learning more tasks does yield increased performance on new, unseen tasks. 
At the same time, the best-performing models often were those that were trained on the evaluation task exclusively (for EC+RC and QA). 
Overall, unsurprisingly, directly learning the task at hand yields the best performing models, but the fact that multiple tasks improve performance for unseen tasks does suggest that the domain-specific knowledge that the PLMs receive in the tuning stage, do help solving the unseen task at hand.
Prompt tuning on the QA and EL (i.e., instruction based finetuning) examples lead to a 3.14\% improvement on the EC + RC task. 
Similarly, prompt tuning on the EC + RC and QA examples yielded a 1.58\% increase in performance on the EL task, with an overall 4.54\% increase over the zero-shot scenario.
A possible explanation, "the ability to recognize whether an entity is an occupation or skill help discriminate whether two entities are not synonymous".
However, training on all tasks did not seem to increase the overall performance on any of the tasks. 
We believe this is potentially caused by overlaps in learned behavior from these different labour market specific tasks and the 1836 tasks FLAN-T5 is already pre-trained on.

\subsection{Implications}
Finetuning PLMs is often an effective transfer mechanism in NLP. 
However, an entire new model is often required for every task. 
Our results indicate that cost-efficient methods such as PTR and instruction-based finetuning can significantly increase the performance of PLMs on downstream labour market applications without introducing any additional model layers, manual annotations, and data augmentation.

Furthermore, our results suggest that while training on general tasks can increase the overall performance on labour market specific applications, providing the general models with labour market specific examples increases performance above and beyond the general finetuning.

\subsection{Limitations}
There are several limitations to the current study that should be considered. First, we only used one-to-one verbalizers between our classes and label words. Meaning that every class label is mapped to one respective label word. This would be a fruitful area for future research, e.g., occupation can also be rewritten as job, or work. Adding these alternatives to the label words will probably yield improved performance over the current one-to-one verbalizers.

Second, for the purpose of this initial exploration we focused primarily on binary classification tasks. As such, we did not incorporate the possibility for a non-existing relation in the PTR finetuning.

Third, while the underlying methods support multiple languages, we chose to conduct our experiments on English. In part because the descriptions used in the QA dataset are not complete for all 28 languages for which ESCO is available. A future study could assess the performance of PTR and instruction based finetuning without examples in other languages.

Lastly, this study primarily focused on the actual \textit{isEssentialFor} and \textit{isOptionalFor} relations as they were present in the ESCO taxonomy. As such, we did not implement the \textit{reversed} and or negative relations, even though this was suggested to further increase performance.

\section{Conclusion}
In this study, we demonstrated that FLAN-T5 substantially outperforms T5 on the QA and EL tasks with respectively 49.7\% and 23.5\% F1 scores. 
However, on the remaining EC+RC task, T5 outperformed FLAN-T5 by 3.53\%. 
Overall it seems that PLMs benefit from instruction based finetuning even on labour market specific benchmarks. 
However, if the task at hand is very different from the task at hand, it can potentially hurt performance, as demonstrated with the EC+RC tasks.

Furthermore, our results seem to indicate that using PTR with labour market specific examples yields improvements above and beyond the 1,836 tasks FLAN-T5 was tuned on.
Unsurprisingly, directly learning the task at hand leads to the best performing models.
But, results also show prompt tuning on other labour market specific tasks can improve performance on unseen tasks.
For example, prompt tuning on EC+RC and QA improved the performance on the EL task with 1.58\%, and prompt tuning on QA and EL improved the performance on the EC+RC task by 3.14\%.

There are several limitations to the current study; i) we solely used one-to-one verbalizer, ii) we focused primarily on binary classification tasks, iii) we only focused on English, and lastly we only used relations actually present in the ESCO taxonomy, meaning that we did not implement the reversed relations. 
Future studies could address the limitations of this study by using incrementing the amount of used label words, adding negative and reversed relations, and using ESCO to construct parallel datasets for all available languages.

\bibliography{bibliography}

\end{document}